\documentclass[11pt]{article}

\usepackage[preprint]{acl}

\usepackage{times}
\usepackage{latexsym}

\usepackage[T1]{fontenc}

\usepackage[utf8]{inputenc}

\usepackage{microtype}

\usepackage{inconsolata}

\usepackage{graphicx}
\usepackage{hyperref}
\usepackage{color}
\usepackage{multirow}
\usepackage{booktabs}
\usepackage{makecell}
\usepackage{graphicx}
\usepackage{amsthm,amsmath,amssymb}
\usepackage{mathrsfs}
\usepackage{pifont}
\usepackage{enumitem}
\usepackage{algorithm}
\usepackage{algorithmic}
\usepackage{array}
\usepackage{tikz} 
\usepackage{tcolorbox}
\usepackage{listings}
\usetikzlibrary{mindmap,trees}
\usepackage{relsize}
\usepackage{subcaption}
\usepackage{amsfonts}
\usepackage{siunitx} 
\usepackage{xcolor}
\usepackage{colortbl}
\usepackage{wrapfig}
\usepackage{float}
\definecolor{masahighlight}{RGB}{232, 245, 233}

\setlist{topsep=2pt,itemsep=1pt,parsep=1pt}

\newcommand{\modelname}{MASA}

\newcommand{\mcsr}{the skill rewriter}

\title{\raisebox{-0.4cm}{
\includegraphics[width=1.1cm]{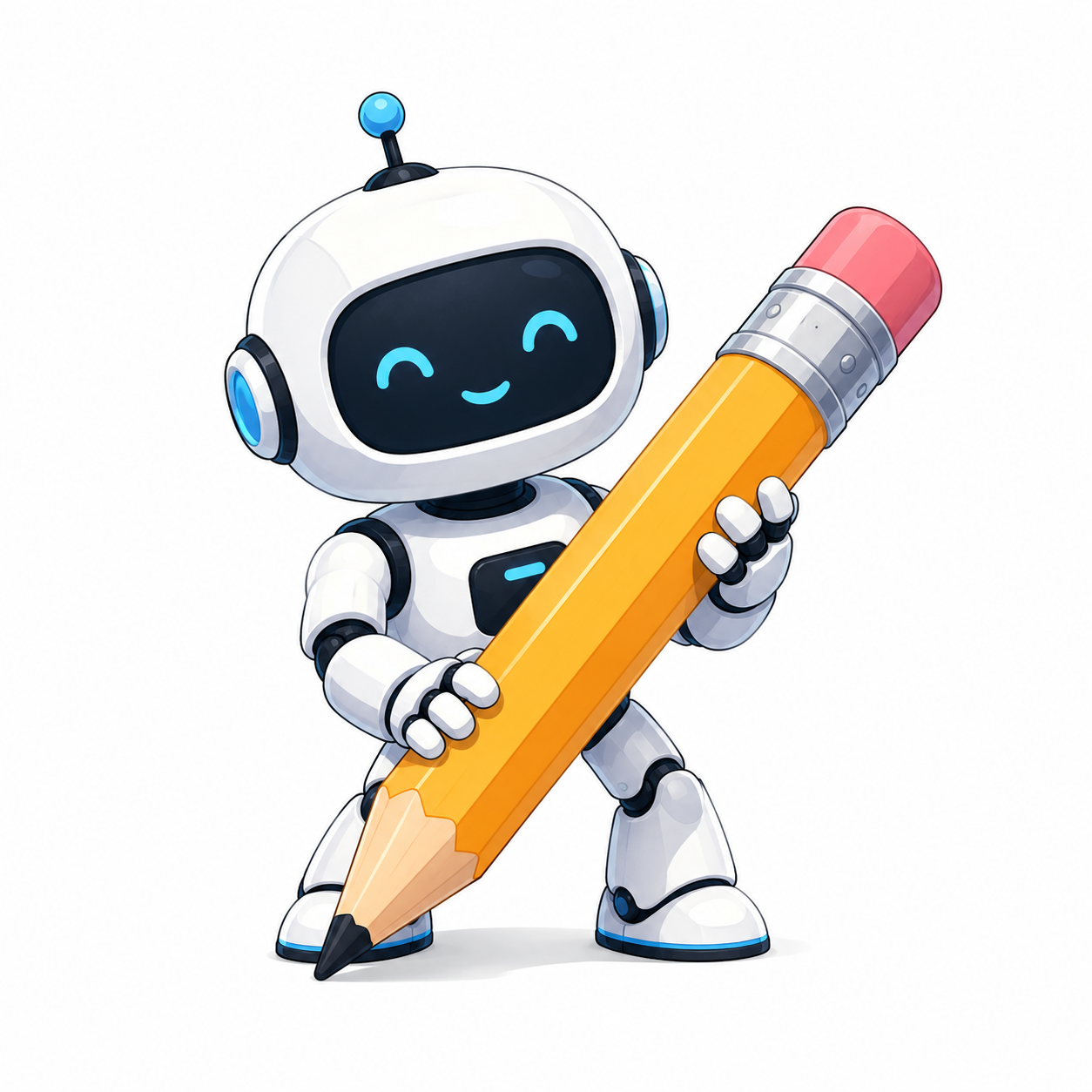}}
Skill is Not One-Size-Fits-All:\\Model-Aware Skill 
Alignment 
for LLM Agents}

\author{Jianxiang Yu,
Jiapeng Zhu,
Bochen Lin,
{\bf Qier Cui,} 
Zichen Ding,
{\bf Xiang Li\thanks{Corresponding author}} \\
East China Normal University,
Shanghai, China \\
\href{mailto:jianxiangyu@stu.ecnu.edu.cn}{jianxiangyu@stu.ecnu.edu.cn}
\\
}

\begin{document}
\maketitle
\begin{abstract}
LLM agents increasingly retrieve externally curated \emph{skills}---procedural instructions retrieved at decision time---to improve performance on long-horizon interactive tasks.
Existing skill libraries are typically treated as model-agnostic, reusing the same skill formulations across backbones with substantially different capacities and behaviors. 
However, our controlled experiments across multiple model scales show that skill effectiveness is strongly model-dependent: a skill that benefits one backbone can harm another.
Motivated by this observation, we propose \modelname{} (\emph{Model-Aware Skill Alignment}), a framework that adapts skills to each target backbone without modifying agent weights.
\modelname{} operates in two stages: (1) a hierarchical skill evolution pipeline that iteratively rewrites general and task-specific skills using hill climbing and UCB-driven tree search, guided by environment feedback and model capability profiles; 
and (2) a lightweight model-conditioned skill rewriter trained on evolution trajectories to reproduce the adaptation in a single forward pass.
Experiments across three interactive environments and four backbones show that \modelname{} consistently achieves the best overall performance, with gains of up to $25.8$ points over the strongest baseline. 
The learned rewriter further generalizes to unseen tasks and environments without additional search, 
consistently outperforming a much larger teacher LLM at a fraction of the inference cost.
Our code is publicly available.\footnote{\url{https://github.com/jianxiangyu/MASA_}}
\end{abstract}

\section{Introduction}
\label{sec:intro}

\begin{figure}[t]
    \centering
    \includegraphics[width=0.99\linewidth]{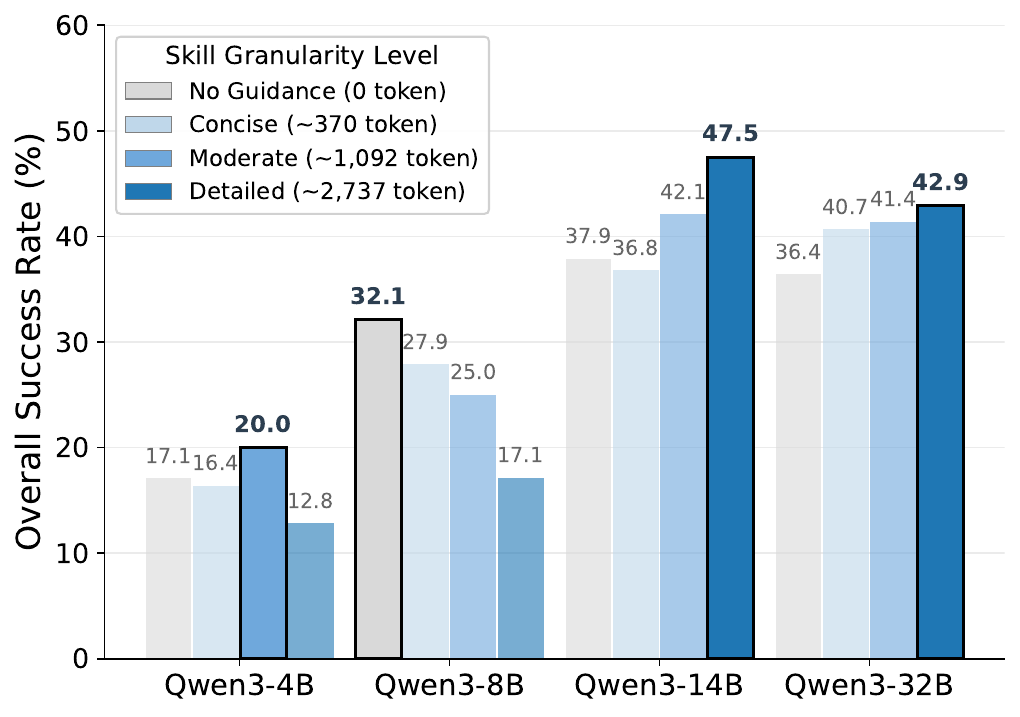}
    \caption{\textbf{Skill granularity is not one-size-fits-all.} ALFWorld success rate (\%) for four Qwen3 backbones under a \texttt{No-Skill} control and three granularity levels (\texttt{Concise}, \texttt{Moderate}, \texttt{Detailed}). The optimal level differs across backbones. 
    }
    \label{fig:pre_qwen}
\end{figure}

LLM agents increasingly solve long-horizon interactive tasks, including web navigation~\cite{ouyang2026skillos}, embodied control~\cite{skill0}, and tool use~\cite{schick2023toolformer,jiang2026sok,wang2024survey,hsiao2025procedural}.
A common approach to steer these agents without modifying model weights is to retrieve short pieces of procedural knowledge---which we call \emph{skills}---from an external library at each step~\cite{wang2023voyager,wang2023jarvis,wang2024agentworkflow,wang2024automanual,zhao2024expel,ma2026skillclaw}.
Existing skill-library systems, whether hand-crafted~\cite{zhu2023ghost} or distilled from agent trajectories~\cite{zhao2024expel,wang2024automanual,xia2026skillrl,wang2025sage}, typically construct a single shared library and reuse it across different LLM backbones.
In practice, deployment constraints such as latency budgets, inference cost, and hardware availability mean that real-world agent systems must operate with backbones of vastly different scales rather than simply relying on the strongest available model~\cite{yao2025efficient,zheng2025review}. 
This deployment heterogeneity raises a critical question for skill-library design: can a single skill formulation serve models with substantially different capacities equally well?

To examine this, we experiment on ALFWorl~\cite{shridhar2020alfworld} (full setup and analysis in \S\ref{sec:prelim}): keeping the principles of a skill library fixed, we vary only its granularity and evaluate four Qwen3 backbones (4B--32B)~\cite{yang2025qwen3}. 
As Figure~\ref{fig:pre_qwen} shows, the optimal granularity varies across models; indeed, a skill that boosts one backbone can actively degrade another. 
A parallel experiment on the Gemma3 family (Appendix~\ref{app:cross_family}) confirms that the same pattern holds across families, and that models of the same size but from different families also prefer different skill formulations.
This observation suggests that the effectiveness of a skill library depends not only on what knowledge it encodes, but also on how that knowledge is expressed relative to the target model's capacity: 
when the expression is misaligned, retrieved skills distract rather than help.
A well-designed skill library should amplify the strengths of its target backbone, 
unlocking capabilities that generic, model-agnostic skills cannot.

We pursue this goal with \textbf{\modelname{}}, \textbf{M}odel-\textbf{A}ware \textbf{S}kill \textbf{A}lignment, a framework that aligns the formulation of a skill library with each target backbone without modifying agent weights.
\modelname{} treats skill alignment as a hierarchical search problem driven by environment feedback.
It first runs a \emph{hierarchical model-conditioned skill evolution}: 
a stronger teacher LLM iteratively rewrites skills guided by a capability profile of the target model, applying hill-climbing over general skills and UCB-driven tree search over task-specific skills.
To eliminate the costly teacher at deployment, the discovered rewrites train a lightweight model-conditioned skill rewriter that adapts skills in a single forward pass, outperforming the teacher while being orders of magnitude cheaper.

Our main contributions are as follows:

\begin{itemize}[leftmargin=*,
itemsep=0pt]
\item We empirically demonstrate that different models require different skill formulations: the same skill library that benefits one backbone can actively degrade another.
This finding challenges the one-size-fits-all assumption and motivates model-aware skill alignment.
\item We propose \modelname{}, 
a framework that aligns skill formulations with each target backbone.
It combines iterative search to evolve optimal skills with a lightweight rewriter that transforms unaligned skills into model-appropriate ones.
\item We evaluate \modelname{} across three diverse environments and four Qwen3 backbones, achieving the highest success rate with gains up to $+25.8$ points.
\modelname{}-rewriter further generalizes to unseen tasks and environments in a single forward pass, outperforming a much larger teacher LLM at negligible cost.
\end{itemize}

\begin{figure*}[t]
    \centering
    \includegraphics[width=1.0\linewidth]{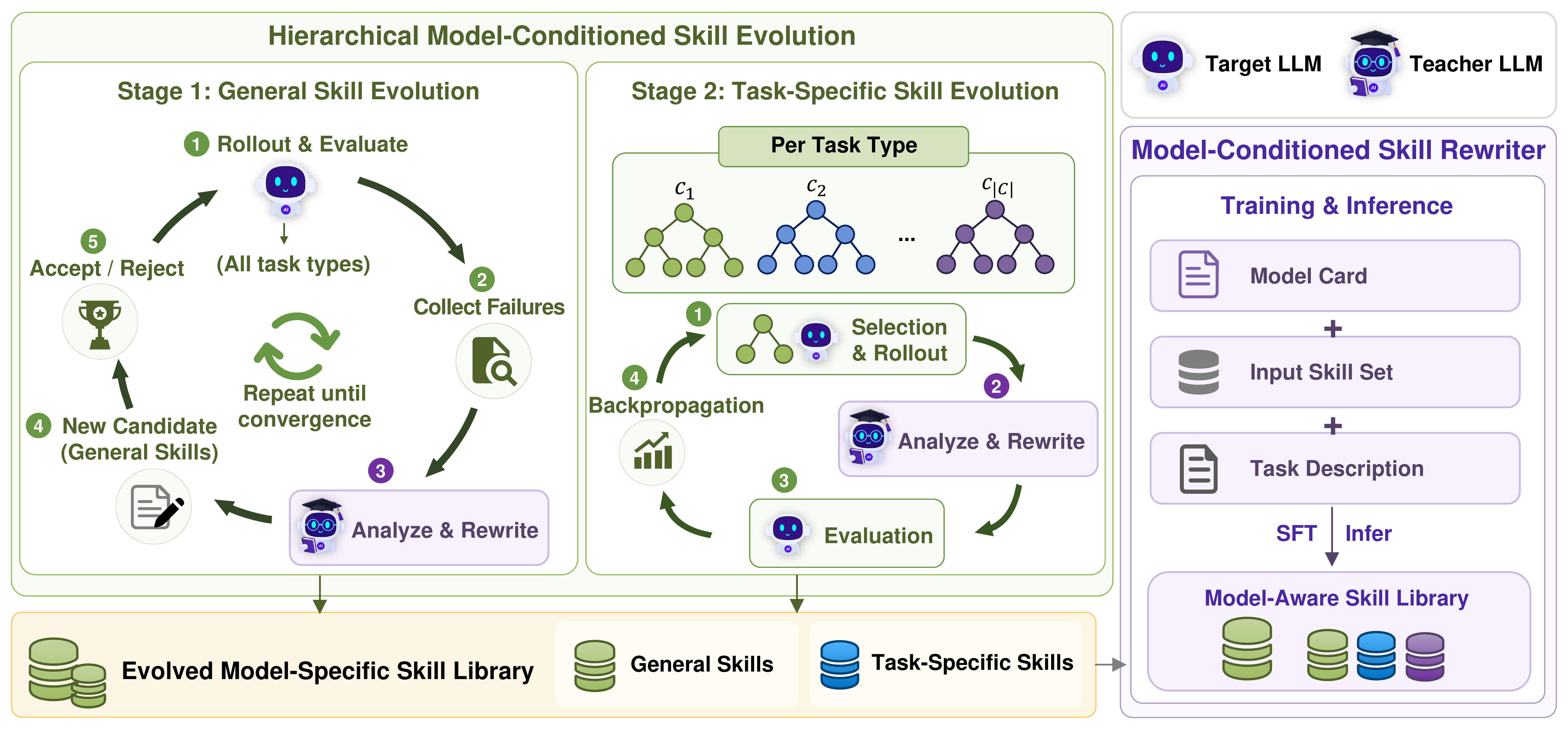}
    \caption{The overall framework of~\modelname{}. 
    }
    \label{fig:framework}
\end{figure*}

\section{Preliminary Study: One Skill Library Does Not Fit All}

\label{sec:prelim}

Before introducing \modelname{}, we ask whether a single skill library serves all model scales equally. To isolate the effect of skill form from skill content, we keep the underlying \emph{principles} fixed and vary only the \emph{granularity} of their textual expression.

\subsection{Setup}
\label{sec:prelim:setup}

We use ALFWorld~\cite{shridhar2020alfworld}, a text-based household task suite spanning six task types, and evaluate on the validation split.
We compare four Qwen3 backbones (4B/8B/14B/32B)~\cite{yang2025qwen3} that differ primarily in capacity while sharing the same architecture and training regimen.
We design one \texttt{No Skill} control and three skill-granularity levels that encode identical behavioral principles but differ in representational depth.
Following prior work, we adopt the skill library of \citet{xia2026skillrl} as the \texttt{Moderate} variant and construct the \texttt{Concise} and \texttt{Detailed} variants through controlled rewriting that preserves the underlying principles while adjusting granularity
(see Table~\ref{tab:app:preliminary_skills} in Appendix~\ref{app:skill_variant_comparison} for side-by-side examples).
All three levels use the same retrieval pipeline, ensuring that observed differences are attributable to granularity alone.

\subsection{Findings}
\label{sec:prelim:findings}

Figure~\ref{fig:pre_qwen} reports overall ALFWorld success rates.

\paragraph{Finding 1: The optimal skill form is model-dependent, and mismatches can hurt.}
No single granularity level is uniformly optimal across models.
Qwen3-4B performs best with \texttt{Moderate skills} while Qwen3-14B and Qwen3-32B achieve their highest scores with \texttt{Detailed skills}.
Notably, Qwen3-8B performs best under the \texttt{No Skill} condition ($32.1\%$) and all three skill variants reduce performance.
Importantly, this does not suggest that skills are inherently incompatible with Qwen3-8B.
Trajectory inspection reveals that, without external guidance, Qwen3-8B often follows short and effective action chains that directly solve the task.
Misaligned skills instead introduce procedural reasoning patterns that override these naturally concise action chains, causing the model to over-explore or deliberate unnecessarily. 
This suggests that the effectiveness of a skill depends not only on its content but also on whether its expression is compatible with the model's default problem-solving strategy.

\paragraph{Finding 2: The granularity--performance relationship is non-monotonic and defies simple heuristics.}
It is unclear how skill granularity should scale with model capability: smaller models may benefit from concise guidance due to limited context utilization capacity, 
yet they may also require more explicit procedural supervision because of weaker reasoning abilities.
Our results show that neither direction holds consistently.
Qwen3-32B underperforms Qwen3-14B by $4.6$ points under \texttt{Detailed} despite being twice the size, inverting the usual scaling trend.
For Qwen3-4B, performance does not monotonically improve in either direction: \texttt{Moderate} outperforms both \texttt{Concise} and \texttt{Detailed}, indicating that the optimum lies at an intermediate level that cannot be reached by simply ``adding more detail'' or ``stripping to minimum.''
This complexity necessitates search-based rather than rule-based skill adaptation.

\paragraph{Finding 3: Performance varies sharply across task types.}
Per-task breakdown (Appendix~\ref{app:prelim_per_task}) reveals that within a given model--granularity pairing, success rates can vary by over $60$ points across task types---a spread far exceeding the differences between granularity levels for any single task. 
For example, Qwen3-14B with \texttt{Concise skills} scores $74.2$ on \textsc{Pick} but only $13.7$ on \textsc{Cool}.
Some task types benefit from detailed skills regardless of model size, while others are insensitive or even harmed. 
This heterogeneity suggests that global optimization alone is insufficient—skill alignment must also operate at the task-type level to address the distinct demands of each task type.

A parallel experiment on the Gemma3 family (4B/12B/27B) reveals the same scale-dependent trend (Appendix~\ref{app:cross_family}), suggesting the phenomenon generalizes across model families.
\paragraph{Implications.}
Together, the three findings impose concrete design requirements:
\begin{enumerate}[leftmargin=1.5em,itemsep=1pt,
label=(\roman*)]
\item \emph{Model-conditioned:} the optimal skill form varies per backbone, so alignment must be explicitly conditioned on the target model's capacity (Finding~1).
\item \emph{Search-based rather than heuristic:} the relationship between skill granularity and performance is non-monotonic and model-specific, ruling out simple alignment rules (Finding~2).
\item \emph{Task-type-specific:} within the same backbone, different task types respond differently to the same skills, 
requiring per-task-type adaptation in addition to global optimization (Finding~3).
\end{enumerate}

We further note that our controlled study varies only one axis of skill form (textual granularity) while holding content fixed.
In practice, misalignment can also arise from differences in decision strategy, framing, or format, suggesting that a complete solution must perform open-ended, model-aware rewriting. 
\modelname{} is designed to address all three requirements.
\section{Method: \modelname{}}
\label{sec:method}

We present \modelname{}, a framework that \emph{conditions} skill evolution on the capability profile of a target backbone, yielding skill libraries specifically adapted to each model rather than relying on a universal, model-agnostic formulation. 
\modelname{} comprises two complementary components: a search-time skill evolution pipeline (Section~\ref{sec:method:rsep}) that evolves skills under explicit capability conditioning provided by a structured \emph{model card}, and a deployment-time skill rewriter (Section~\ref{sec:method:mcsr}) that learns this model-conditioned rewriting policy and adapts new skills in a single forward pass.
An overview of the framework is shown in Figure~\ref{fig:framework}.

\subsection{Problem Formulation and Skill Library}
\label{sec:method:formulation}

\paragraph{Agent setup.}
A frozen LLM agent $F$ interacts with environment $\mathcal{E}$. At each step $t$, the agent receives observation $o_t$, retrieves relevant skills from a skill library $\mathcal{S}$, and produces an action $a_t$ 
:
\begin{equation}
\label{eq:agent}
a_t \sim F\!\left(\cdot \mid \tau_{<t},\; \hat{\mathcal{S}}_t\right), \quad \hat{\mathcal{S}}_t = \mathrm{TopK}(\mathcal{S}, o_t, k),
\end{equation}
where $\tau_{<t} = (o_1, a_1, \ldots, o_{t-1}, a_{t-1})$ is the interaction history and $\mathrm{TopK}$ retrieves the $k$ most relevant skills by cosine similarity. 
The backbone $F$ remains frozen throughout, and the sole optimization variable is the skill library $\mathcal{S}$.

\paragraph{Hierarchical skill library.}
Following~\citet{xia2026skillrl}, we structure $\mathcal{S}$ into two levels: \emph{general skills} $\mathcal{S}^{G}$ (cross-task strategy principles) and \emph{task-specific skills} $\mathcal{S}^{T} = \{\mathcal{S}^{T_c}\}_{c \in \mathcal{C}}$, where each $\mathcal{S}^{T_c}$ contains action guidelines tailored to task type $c$ and $\mathcal{C}$ is the set of all task types. 
At inference, a lightweight encoder (Qwen3-Embedding-0.6B) separately retrieves top-$k_G$ general skills and top-$k_T$ task-specific skills for the current observation.

\paragraph{Model card.}
The key conditioning signal in \modelname{} is the \emph{model card} $\mathcal{M}_F$, a structured profile of a target backbone $F$. Each card contains: 
(i)~\emph{architecture metadata} (model family, parameter count, layer/attention configuration, context window), 
(ii)~\emph{training provenance} 
(training data scale, multilingual support), 
and (iii)~\emph{capability profile} (strengths and weaknesses of the backbone). The construction protocol is detailed in Appendix~\ref{app:model_cards}.

\paragraph{Objective.}
We define a per-episode adjusted reward $R$ that balances task completion against skill-induced stalling:
\begin{equation}
\label{eq:reward}
R(F, \mathcal{S}, e) = \mathrm{SR}(F,\mathcal{S},e) - \lambda \cdot \mathrm{NHR}(F,\mathcal{S},e),
\end{equation}
where $e$ denotes a single episode, $\mathrm{SR} \in \{0,1\}$ is task success, 
$\mathrm{NHR}$ is the \emph{nothing-happens rate}---the fraction of steps after which the environment state remains unchanged, serving as a proxy for skill-induced stalling (e.g., the agent repeatedly issuing ineffective or invalid actions),
and $\lambda \in [0,1]$ controls the penalty strength.
The overall optimization objective seeks the skill library maximizing expected adjusted reward over a set of evaluation episodes $\mathcal{D}$:
\begin{equation}
\label{eq:obj}
\mathcal{S}^\star_F = \arg\!\max_{\mathcal{S}}\; \mathbb{E}_{e \sim \mathcal{D}}\!\left[ R(F, \mathcal{S}, e) \right],
\end{equation}
where $\mathcal{S}^\star_F$ denotes the optimal skill library adapted to backbone $F$.

\subsection{Hierarchical Model-Conditioned Skill Evolution}
\label{sec:method:rsep}

The skill evolution pipeline is a teacher-driven search over skill texts. A stronger \emph{teacher} LLM $T$ (i) analyzes failure trajectories of $F$ to produce a structured failure attribution and (ii) rewrites skills conditioned on the model card $\mathcal{M}_F$, steering edits toward formulations compatible with $F$'s observed characteristics.

\paragraph{Two-stage optimization.}
The pipeline optimizes $\mathcal{S}^{G}$ and $\{\mathcal{S}^{T_c}\}$ in separate stages, motivated by both computational efficiency and conceptual separation.
From a computational perspective, a single edit to $\mathcal{S}^{G}$ requires evaluation over the full task suite, whereas edits to $\mathcal{S}^{T_c}$ affect only a single task type.
From a modeling perspective, the two skill levels encode fundamentally different forms of knowledge.
General skills capture backbone-level behavioral guidance that is intended to transfer across tasks (e.g., ``always verify your action parsed correctly''), while task-specific skills encode domain procedures tailored to particular environments (e.g., ``check the fridge before the counter'').
Therefore,
separating the two stages simplifies credit assignment across the two skill levels while substantially reducing search cost.

\paragraph{Stage 1: General skills via iterative hill climbing.}

General skills encode high-level behavioral priors that affect agent behavior across many task types.
Evaluating a candidate general skill requires running the agent across the full task suite and aggregating feedback over diverse environments, making exhaustive search prohibitively expensive.
We therefore optimize $\mathcal{S}^{G}$ via iterative hill climbing~\cite{russell2010aima}, which provides a simple and effective strategy for progressively improving the current skill set under environment feedback.

Each iteration proceeds as follows.
\emph{Rollout}: the target model $F$ equipped with the current best general skills is rolled out across all task types to compute the  reward across episodes.
\emph{Analysis}: the teacher collects failed trajectories from these rollouts and produces a structured failure attribution focusing on general behavioral deficiencies rather than task-specific procedural gaps.
\emph{Rewrite}: given the current best skill set, the failure attribution, the model card $\mathcal{M}_F$, and the $K$ highest-reward skill sets from all previous iterations (which help the teacher learn from the full optimization trajectory rather than only the most recent failure), the teacher outputs a revised general skill set.
\emph{Accept/Reject}: the new candidate is evaluated on the full task suite and accepted only if it achieves a higher reward than the current best.
Search terminates after at most $I$ iterations or after $p$ consecutive iterations without improvement.

\paragraph{Stage 2: Task-specific skills via per-type tree search.}
Unlike general skills, task-specific skills encode domain procedures where multiple structurally different strategies may be effective for the same task type. 
This motivates a tree-structured search that can explore diverse branches rather than committing to a single refinement path.
We run an independent tree search per task type $c$, where each node holds a candidate task-specific skill set $\mathcal{S}^{T_c}$ and each edge corresponds to a teacher rewrite.

Each iteration proceeds in four steps.
\emph{Selection}: starting from the root, UCB1~\cite{kocsis2006bandit} is applied recursively to select the most promising leaf node, balancing exploitation of high-reward nodes with exploration of less-visited ones.
\emph{Expansion}: the target model $F$ is rolled out on type-$c$ tasks using the selected node's skill set, and the teacher collects failed trajectories, produces a failure attribution, and outputs a revised task-specific skill set---forming a new child node.
\emph{Evaluation}: the new child's skill set is evaluated on type-$c$ tasks to obtain its average reward.
\emph{Backpropagation}: the reward is propagated from the new node back to the root, updating visit counts and value estimates along the path.
Per-type trees are independent and fully parallelizable.

Overall,
the two stages run sequentially: $\mathcal{S}^{G\star}_F$ obtained in Stage~1 is held fixed throughout Stage~2, and the final output is a \emph{model-specific} skill library $\mathcal{S}^\star_F = (\mathcal{S}^{G\star}_F, \{\mathcal{S}^{T_c\star}_F\}_{c \in \mathcal{C}})$.
The detailed procedures are given in Algorithms~\ref{alg:stage1} and~\ref{alg:stage2}, and further details of the two-stage search 
are provided in Appendix~\ref{app:hyperparams}.

\subsection{Model-Conditioned Skill Rewriter}
\label{sec:method:mcsr}

The skill evolution pipeline delivers strong skill libraries but requires substantial compute (hundreds to thousands of full-environment rollouts) and an environment-provided reward signal. 
\modelname{}-Rewriter addresses this by learning the rewriting policy that the evolution pipeline implicitly executes, enabling cheap adaptation to new domains and tasks without further environment interaction.

\paragraph{Training data.}
Each training instance maps an input skill set to the corresponding evolved optimum:
\begin{equation}
\label{eq:mcsr_data}
(\mathcal{M}_F,\; \mathcal{S}_{F_{\text{in}}},\; d) \longrightarrow \mathcal{S}^\star_F,
\end{equation}
where $\mathcal{M}_F$ is the model card, $d$ is the task description, and $\mathcal{S}_{F_{\text{in}}}$ is an input skill set (either general $\mathcal{S}^G$ or task-specific $\mathcal{S}^{T_c}$).
The output $\mathcal{S}^\star_F$ is always drawn from the evolution pipeline's high-scoring skill sets.
To ensure the rewriter learns to improve skills regardless of their initial quality, $\mathcal{S}_{F_{\text{in}}}$ is deliberately sampled from sources spanning a wide quality range:
(i)~\emph{search intermediates} at early, mid, and late stages of the evolution pipeline;
(ii)~\emph{cross-model transfers}---optimal skills from a different backbone;
(iii)~\emph{one-shot teacher rewrites} without iterative search;
and (iv)~\emph{augmented variants} (noisy, partial, or verbose perturbations of existing skills).
This diversity exposes the rewriter to the full range of input conditions it may encounter at deployment.
Additional details are provided in Appendix~\ref{app:mcsr_train}.

\paragraph{Training.}
We instantiate \mcsr{} with Qwen3-4B, a lightweight backbone chosen to keep deployment cost minimal while retaining sufficient capacity for structured rewriting. The model is trained via supervised fine-tuning (SFT) with cross-entropy loss:
\begin{equation}
\label{eq:mcsr_loss}
\mathcal{L} = -\mathbb{E}_{\mathcal{D}_{\text{train}}} \!\left[ \log\, p_\theta\!\left(\mathcal{S}^\star_F \;\middle|\; \mathcal{M}_F,\, \mathcal{S}_{F_{\text{in}}},\, d\right) \right].
\end{equation}

\paragraph{Inference.}
At deployment, given the target backbone's model card $\mathcal{M}_F$, an input skill set $\mathcal{S}_{F_{\text{in}}}$, and the task description $d$, \mcsr{} produces an adapted skill set in a single forward pass:
\begin{equation}
\label{eq:mcsr_infer}
\mathcal{S}^\star_{F} = f_\theta\!\left(\mathcal{M}_{F},\; \mathcal{S}_{F_{\text{in}}},\; d\right),
\end{equation}
without requiring any environment interaction or iterative search.

\begin{table*}[t]
\centering
\small
\setlength{\tabcolsep}{4pt}
\renewcommand{\arraystretch}{1.08}
\resizebox{0.965\textwidth}{!}{%
\begin{tabular}{ll rrrrrr c >{\color{gray!80!black}}c | rr >{\color{gray!70!black}}r}
\toprule
& & \multicolumn{8}{c|}{\textbf{ALFWorld}} & \multicolumn{3}{c}{\textbf{WebShop}} \\
\cmidrule(lr){3-10}\cmidrule(lr){11-13}
\textbf{Model} & \textbf{Method}
& \textbf{Pick}
& \textbf{Look}
& \textbf{Clean}
& \textbf{Heat}
& \textbf{Cool}
& \textbf{Pick2}
& \textbf{SR $\uparrow$}
& \textbf{Steps $\downarrow$}
& \textbf{SR $\uparrow$}
& \textbf{Score $\uparrow$}
& \textbf{Steps $\downarrow$} \\
\midrule

\multirow{4}{*}{Qwen3-4B}
& No Skill              & 20.0 & 15.4 & 18.5 & 18.8 & 16.0 & 12.5 & 17.1 & 44.6 & 23.0 & 42.2 & 9.5 \\
& + Base Skill          & 20.0 & 30.8 & 29.6 & 12.5 & 20.0 &  8.3 & 20.0 & 42.3 & 19.4 & 34.8 & 11.0 \\
& + DS-Adapter          & \textbf{28.6} & 30.8 & 37.0 & 18.8 & \textbf{32.0} & 12.5 & 27.1 & 40.0 & 19.2 & 25.7 & 12.4 \\
\rowcolor{masahighlight}
& + \modelname{}        & 25.7 & \textbf{53.8} & \textbf{40.7} & \textbf{37.5} & 24.0 & \textbf{20.8} & \textbf{31.4} & \textbf{38.4} & \textbf{26.4} & \textbf{49.1} & \textbf{8.4} \\
\midrule

\multirow{4}{*}{Qwen3-8B}
& No Skill              & 54.3 & \textbf{46.2} & 29.6 &  6.2 & 24.0 & 20.8 & 32.1 & 39.1 & 4.6 & 32.7 & 10.0 \\
& + Base Skill          & 17.1 & 38.5 & 40.7 & 31.2 & 20.0 & 12.5 & 25.0 & 40.5 & 6.0 & 32.6 & 9.3 \\
& + DS-Adapter          & 25.7 & 38.5 & 44.4 & 25.0 & 16.0 & 16.7 & 27.1 & 39.6 & 4.4 & 18.2 & 12.6 \\
\rowcolor{masahighlight}
& + \modelname{}        & \textbf{62.9} & 38.5 & \textbf{70.4} & \textbf{75.0} & \textbf{56.0} & \textbf{37.5} & \textbf{57.9} & \textbf{29.2} & \textbf{28.6} & \textbf{60.1} & \textbf{4.7} \\
\midrule

\multirow{4}{*}{Qwen3-14B}
& No Skill              & 65.7 & 38.5 & 25.9 & 43.8 & 16.0 & 29.2 & 37.9 & 36.7 & 2.8 & 19.9 & 12.7 \\
& + Base Skill          & 68.6 & 46.2 & 44.4 & 25.0 & 20.0 & 33.3 & 42.1 & 34.1 & 1.6 & 14.8 & 13.5 \\
& + DS-Adapter          & 68.6 & \textbf{53.8} & 40.7 & 18.8 & 40.0 & 29.2 & 44.3 & 34.8 & 2.0 & 12.6 & 13.6 \\
\rowcolor{masahighlight}
& + \modelname{}        & \textbf{85.7} & \textbf{53.8} & \textbf{81.5} & \textbf{56.2} & \textbf{44.0} & \textbf{45.8} & \textbf{64.3} & \textbf{25.7} & \textbf{29.2} & \textbf{54.4} & \textbf{8.0} \\
\midrule

\multirow{4}{*}{Qwen3-32B}
& No Skill              & 48.6 & \textbf{46.2} & 44.4 & 25.0 & 32.0 & 16.7 & 36.4 & 37.0 & 6.6 & 35.2 & 9.9 \\
& + Base Skill          & 48.6 & \textbf{46.2} & 40.7 & 50.0 & 44.0 & 20.8 & 41.4 & 35.6 & 7.2 & 24.2 & 12.0 \\
& + DS-Adapter          & 51.4 & 38.5 & 59.3 & 37.5 & 44.0 & 29.2 & 45.0 & 32.2 & 3.6 & 14.3 & 13.3 \\
\rowcolor{masahighlight}
& + \modelname{}        & \textbf{57.1} & \textbf{46.2} & \textbf{77.8} & \textbf{81.3} & \textbf{76.0} & \textbf{54.2} & \textbf{65.7} & \textbf{24.3} & \textbf{34.6} & \textbf{59.9} & \textbf{7.3} \\
\bottomrule
\end{tabular}%
}
\caption{Performance on ALFWorld and WebShop. ALFWorld reports per-task and average success rate (SR~\%), and average interaction steps across all task types; WebShop reports average SR~(\%), score, and average steps. \textbf{Bold} marks the best within each backbone.}
\label{tab:main:alfworld}
\end{table*}

\paragraph{Complementary roles.}
The skill evolution pipeline provides per-backbone upper bounds via explicit search and produces \mcsr{}'s training signal for \mcsr{}. 
\modelname{}-Rewriter amortizes this search into a single forward pass, enabling rapid adaptation without environment interaction. 
The evolution pipeline is preferred when rollout budget permits thorough optimization, whereas \mcsr{} is better suited to compute-constrained deployment scenarios.

\section{Experiments}
\label{sec:experiment}

We evaluate whether model-conditioned skill evolution outperforms model-agnostic baselines across diverse environments and backbones, and whether \modelname{}-Rewriter can generalize the learned rewriting policy to held-out task types and unseen environments without additional search.

\subsection{Experimental Setup}
\label{sec:exp:setup}

\paragraph{Environments.} We evaluate on three environments spanning distinct action spaces and reasoning demands.
(i)~ALFWorld~\cite{shridhar2020alfworld} is a text-based embodied environment where agents complete household tasks (e.g., heating, cleaning, picking up objects) by issuing text commands to navigate rooms and interact with objects. It contains six task types with varying difficulty.
(ii)~WebShop~\cite{yao2022webshop} simulates online shopping: agents navigate a realistic web interface, search for products, compare attributes, and make purchase decisions that satisfy natural-language user specifications.
(iii)~Search-augmented QA requires agents to retrieve and synthesize information from web search results. 
We include seven benchmarks covering both single-hop (NQ~\cite{kwiatkowski2019nq}, TriviaQA~\cite{joshi2017triviaqa}, PopQA~\cite{mallen2023popqa}) and multi-hop reasoning (HotpotQA~\cite{yang2018hotpotqa}, 2Wiki~\cite{ho20202wiki}, MuSiQue~\cite{trivedi2022musique}, Bamboogle~\cite{press2023bamboogle}).

\paragraph{Backbones and baselines.} 
Target agents are Qwen3-\{4B, 8B, 14B, 32B\}~\cite{yang2025qwen3};
\footnote{All Qwen3 backbones are used in non-thinking mode. 
This choice reflects typical deployment scenarios where latency and token budgets are constrained, and ensures that observed performance differences are attributable to skill design rather than reasoning-mode configuration.} 
the teacher LLM is DeepSeek-V4-Pro~\cite{deepseek2026v4}. 
We compare against three baselines:
(1) \emph{No Skill} (the raw backbone without any skill augmentation),
(2) \emph{Base Skill} (the initial skill library from SkillRL~\cite{xia2026skillrl}, shared across all backbones without model-specific adaptation), and
(3) \emph{DS-Adapter} (a one-shot teacher rewrite that adapts the Base Skill library conditioned on the model card $\mathcal{M}_F$, without iterative search).

\noindent The Base Skill library also serves as the initialization $\mathcal{S}^{G_0}_F$ and $\mathcal{S}^{T_{c_0}}_F$ for \modelname{}'s evolution pipeline.

\subsection{Skill Evolution Evaluation}
\label{sec:exp:main}

\begin{table*}[t]
\centering
\small
\setlength{\tabcolsep}{4pt}
\renewcommand{\arraystretch}{1.08}
\resizebox{0.95\textwidth}{!}{%
\begin{tabular}{ll ccc | cccc | c}
\toprule
\multirow{2}{*}{\textbf{Model}} & \multirow{2}{*}{\textbf{Method}} & \multicolumn{3}{c|}{\textbf{Single-Hop QA}} & \multicolumn{4}{c|}{\textbf{Multi-Hop QA}} & \multirow{2}{*}{\textbf{Avg.}} \\
\cmidrule(lr){3-5}\cmidrule(lr){6-9}
& & \textbf{NQ}$^{\dagger}$ & \textbf{TriviaQA}$^{\star}$ & \textbf{PopQA}$^{\star}$ & \textbf{HotpotQA}$^{\dagger}$ & \textbf{2Wiki}$^{\star}$ & \textbf{MuSiQue}$^{\star}$ & \textbf{Bamboogle}$^{\star}$ & \\
\midrule
\multirow{4}{*}{Qwen3-4B}
& No Skill      & 29.4 & 51.0 & 37.2 & 27.7 & 22.8 & 6.4 &  9.3 & 32.9 \\
& + Base Skill   & 34.5 & \textbf{57.4} & 38.2 & 28.5 & 24.4 & 7.8 & 10.1 & 35.5 \\
& + DS-Adapter   & 33.0 & 56.5 & \textbf{41.8} & \textbf{28.6} & 23.9 & 9.3 & 12.9 & 36.2 \\
\rowcolor{masahighlight}
& + \modelname{} & \textbf{35.5} & 55.3 & 38.9 & 27.4 & \textbf{27.0} & \textbf{9.4} & \textbf{61.3} & \textbf{36.9} \\
\midrule
\multirow{4}{*}{Qwen3-8B}
& No Skill      & 19.1 & 46.5 & 30.3 & 24.8 & \textbf{30.6} & 6.7 & \textbf{68.1} & 31.3 \\
& + Base Skill   & 34.0 & \textbf{58.5} & 38.8 & \textbf{28.6} & 25.9 & 6.2 & 10.1 & 36.1 \\
& + DS-Adapter   & 33.2 & 57.6 & 38.7 & 27.8 & 22.9 & 5.6 &  7.7 & 35.0 \\
\rowcolor{masahighlight}
& + \modelname{} & \textbf{36.4} & 56.7 & \textbf{39.0} & \textbf{28.6} & 25.7 & \textbf{10.0} & 62.5 & \textbf{37.2} \\
\midrule
\multirow{4}{*}{Qwen3-14B}
& No Skill      & 33.8 & 60.2 & 40.6 & 31.7 & 26.8 &  7.6 & 10.5 & 37.6 \\
& + Base Skill   & 35.3 & 60.5 & 39.5 & 32.7 & \textbf{30.3} & \textbf{11.4} & \textbf{15.3} & 38.5 \\
& + DS-Adapter   & 33.9 & 60.2 & 39.5 & 31.6 & 28.5 &  9.2 & 12.5 & 37.7 \\
\rowcolor{masahighlight}
& + \modelname{} & \textbf{35.6} & \textbf{61.8} & \textbf{40.7} & \textbf{32.8} & 30.0 & 9.7 & 8.9 & \textbf{39.0} \\
\midrule
\multirow{4}{*}{Qwen3-32B}
& No Skill      & 29.1 & 59.8 & 38.3 & 32.2 & 29.3 &  8.6 & 64.5 & 38.1 \\
& + Base Skill   & 33.8 & 61.4 & 39.3 & 33.8 & 26.0 & 11.7 & \textbf{67.7} & 38.7 \\
& + DS-Adapter   & 34.4 & 61.5 & \textbf{40.6} & 34.0 & 32.0 & 11.6 & 64.1 & 40.6 \\
\rowcolor{masahighlight}
& + \modelname{} & \textbf{37.0} & \textbf{61.6} & 40.0 & \textbf{34.2} & \textbf{35.6} & \textbf{11.8} & 66.1 & \textbf{41.5} \\
\bottomrule
\end{tabular}%
}
\caption{Search-augmented QA results (success rate~\%). Skill evolution is conducted on NQ and HotpotQA; $\dag$ and $\star$ indicate in-domain and out-of-domain datasets, respectively.  \textbf{Bold} marks the best within each backbone.}
\label{tab:main:search}
\end{table*}

Table~\ref{tab:main:alfworld} reports ALFWorld and WebShop results across all four backbones.

\paragraph{ALFWorld.}
\modelname{} achieves the highest average success rate for every backbone: $31.4$ (4B), $57.9$ (8B), $64.3$ (14B), and $65.7$ (32B), with gains of $+4.3$, $+25.8$, $+20.0$, and $+20.7$ over the strongest baseline respectively.
We highlight several observations:

\noindent\textbf{(1) Per-task dominance.}
Beyond the aggregate, \modelname{} achieves the best per-task SR in most individual task types. For Qwen3-14B and 32B, \modelname{} ranks first on \emph{all six} task types simultaneously, 
indicating that the evolved skills improve overall performance without sacrificing coverage across tasks.

\noindent\textbf{(2) Model-agnostic skills can hurt.}
Base Skill and DS-Adapter exhibit severe performance drops on individual tasks, indicating that generic or one-shot adapted skills can introduce model-specific conflicts.
In contrast,
\modelname{} 
avoids these regressions through iterative model-conditioned search.

\noindent\textbf{(3) Scaling behavior.}
For 8B and above, the backbones have sufficient capacity to leverage model-specific skills, yielding substantial improvements.
The gain on 4B is comparatively modest, likely due to the backbone's inherent capability ceiling limiting how much skill guidance can help.

\noindent\textbf{(4) Inference efficiency.}
\modelname{} consistently reduces average interaction steps (e.g., 8B: $39.1 \to 29.2$; 14B: $36.7 \to 25.7$). By tailoring skills to each backbone's specific behavior patterns, \modelname{} helps the agent locate target objects and execute correct action sequences more precisely, reducing redundant exploration and failed attempts.

\paragraph{WebShop.}
\modelname{} again achieves the highest success rate and score for every backbone, substantially outperforming all baselines.
WebShop reveals a critical challenge for larger Qwen3 models:

\noindent\textbf{(1) Larger models perform worse than 4B without adaptation.}
Notably, 8B/14B/32B baselines all underperform 4B on WebShop (e.g., 14B No Skill: $2.8\%$ vs.\ 4B No Skill: $23.0\%$).
We trace this to excessive chain-of-thought generation: larger models produce verbose reasoning preambles before each action, inflating action length and exhausting the step budget on deliberation rather than environment interaction (detailed statistics in Appendix~\ref{app:webshop_trajectory}).
Since model-agnostic skills are not designed to address this model-specific behavioral pattern, they provide limited benefit and in some cases further degrade performance.

\noindent\textbf{(2) \modelname{} addresses this challenge.}
By evolving skills conditioned on each backbone's behavioral profile, \modelname{} guides models toward effective action patterns---achieving SR of $26.4$ (4B), $28.6$ (8B), $29.2$ (14B), and $34.6$ (32B), far surpassing all baselines.
The efficiency gain is also notable: baselines that do succeed average $12$--$13$ steps, whereas \modelname{} achieves higher SR in only $7$--$8$ steps,
dropping to just $4.7$ steps on 8B.

\paragraph{Search-augmented QA.}
Table~\ref{tab:main:search} shows that \modelname{} achieves the highest average SR for every backbone.
Skill evolution is conducted only on NQ and HotpotQA, yet the gains generalize strongly to out-of-domain benchmarks ($\star$)---e.g.,
on 4B,
\modelname{} improves Bamboogle from $12.9$ (best baseline) to $61.3$.
On the largest backbone (32B), \modelname{} ranks first on 5 out of 7 datasets.
These results demonstrate that the evolved skills capture transferable strategies for retrieval and reasoning, rather than overfitting to the datasets used during skill evolution.

\subsection{Skill Rewriter Generalization}
\label{sec:exp:mcsr_ood}

\begin{figure}[t]
    \centering
    \includegraphics[width=\columnwidth]{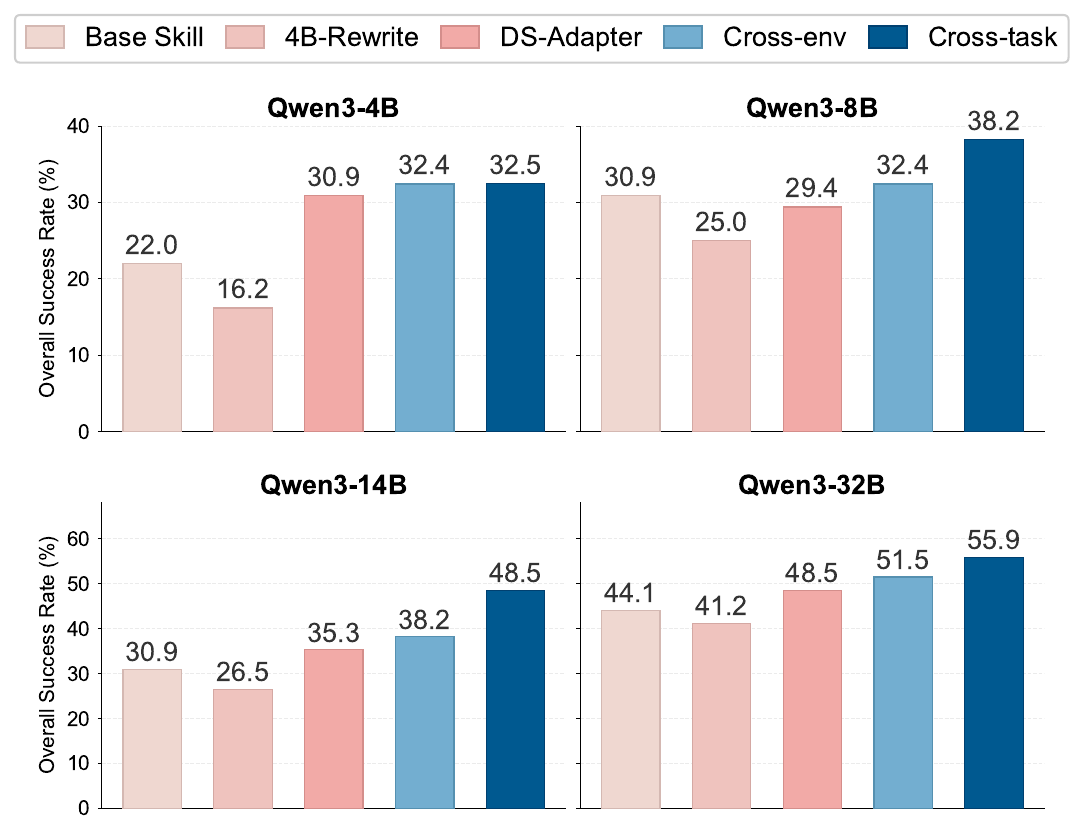}
    \caption{OOD generalization of \modelname{}-Rewriter on held-out ALFWorld task types. 
\textcolor{pink}{Pink} bars denote baselines and
  \textcolor{blue!50!black}{blue} bars denote \modelname{}-Rewriter variants.
    }
    \label{fig:mcsr_ood}
\end{figure}

We evaluate whether \modelname{}-Rewriter can adapt skills for task types not seen during its training, by holding out three ALFWorld task types (\textsc{Clean}, \textsc{Heat}, \textsc{Cool}) and asking \modelname{}-Rewriter to produce task-specific skills for these types.
The general skills remain unchanged.

We compare against three baselines: \emph{Base Skill}, \emph{4B-Rewrite} (Qwen3-4B used as a rewriter without SFT---i.e., the same architecture as \modelname{}-Rewriter but without learned rewriting ability), and \emph{DS-Adapter} (one-shot teacher rewrite targeting the specific held-out task). The two \modelname{}-Rewriter variants differ in training data composition:

\paragraph{Cross-environment transfer.}
\modelname{}-Rewriter is trained exclusively on skill evolution traces from Search and WebShop, then applied to ALFWorld without any in-environment data.
Despite the substantial environment gap (different action spaces and observation formats), Cross-env \modelname{}-Rewriter outperforms DS-Adapter on all four backbones (Figure~\ref{fig:mcsr_ood}), with gains of $+1.5$ (4B), $+3.0$ (8B), $+2.9$ (14B), and $+3.0$ (32B).
This demonstrates that the learned rewriting policy captures model-specific adaptation patterns that transfer across environments---the rewriter can produce useful skills even without exposure to the target environment during training.

\paragraph{Cross-task transfer.}
Building on Cross-env, this variant additionally uses evolution traces from three other ALFWorld task types (\textsc{Pick}, \textsc{Look}, \textsc{Pick2}), excluding the held-out evaluation types.
Cross-task \modelname{}-Rewriter achieves substantially larger gains: $+8.8$ (8B), $+13.2$ (14B), and $+7.4$ (32B) over DS-Adapter.
The gains over Cross-env are especially large on 8B ($+5.8$) and 14B ($+10.3$), suggesting that even skills from unrelated task types provide valuable supervision for adapting to environment-specific interaction patterns and observation structures.

Notably, \modelname{}-Rewriter (4B parameters) consistently surpasses DS-Adapter powered by DeepSeek-V4, demonstrating that a small trained rewriter can outperform a much larger teacher at a fraction of the inference cost.

\textit{Due to space constraints, additional materials including extended related work discussion, ablations, supplementary validation on Gemma3, qualitative examples, and full hyperparameter details are reported in Appendices~\ref{sec:related}--\ref{app:analysis}.
}

\section{Conclusion}
\label{sec:conclusion}

We presented \modelname{}, motivated by the observation that the one-size-fits-all assumption in agent skill design breaks down across model scales.
\modelname{} addressed this through hierarchical skill evolution and a lightweight model-conditioned rewriter that amortizes search into a single forward pass. 
Across three environments and four Qwen3 backbones, 
\modelname{} achieved the best success rate in all settings, with gains up to $+25.8$ points. 
The rewriter further generalized to unseen tasks and environments at negligible deployment cost.

We hope this work motivates treating skills as model-aware artifacts that should be adapted to their target backbone rather than shared uniformly across models of different capacities. 
With proper alignment, even compact models can exhibit behaviors traditionally associated with frontier-scale systems, enabling more accessible and resource-efficient deployment. 
Looking forward, we envision \modelname{}-Rewriter as a lightweight plug-and-play middleware that automatically rewrites existing skill libraries for new backbones, requiring no environment rollouts, retraining, or manual prompt engineering. 
This positions skill alignment as infrastructure rather than a per-deployment engineering effort.

\section*{Limitations}
\label{sec:limitations}

Our empirical evidence is currently restricted to the Qwen3 family
  (4B/8B/14B/32B); extending the skill evolution and rewriter to cover
  more model families---both open-weight (e.g., Llama~\cite{grattafiori2024llama},
  Mistral~\cite{liu2026ministral}) and proprietary
  (e.g., GPT-o3~\cite{openai2025o3}, Claude~\cite{anthropic2024claude3})---and more diverse environments would further strengthen the generality of
  \mcsr{}, though it requires substantially more compute.
In particular, applying the evolution pipeline to closed-source models
  demands hundreds of environment rollouts through paid APIs, making the per-backbone search cost significantly higher than for locally hosted
  models; the \mcsr{} rewriter offers a partial remedy by amortizing
  this cost once trajectories from a few backbones are available.

Additionally, \mcsr{} is trained on skill-evolution trajectories collected from ALFWorld, WebShop, and Search-QA, and the evolution pipeline itself relies on environments that provide automatic success/failure signals (e.g., task completion flags) to judge whether a rewritten skill is effective.
Incorporating domains without such built-in reward signals (e.g., open-ended web tasks or real-world applications~\cite{sun2025genesis}) would require designing external evaluators or human annotations, but would enable the framework to serve an even broader range of agent applications.

\section*{Ethical Considerations}
\label{sec:ethics}

\textbf{Data and Licensing.} \modelname{} does not introduce new data collection from human subjects; all experiments use standard public benchmarks (ALFWorld, WebShop, and open-domain QA datasets) and publicly released models accessed in accordance with their respective licenses.

\textbf{Safety of Agent Empowerment.} By improving the effectiveness of LLM agents through skill adaptation, 
\modelname{} may also increase the capability of agents operating in interactive environments. 
Overall, the framework should be deployed in safety-critical or high-risk settings with additional monitoring, policy constraints, and human oversight.

\textbf{Bias and Reliability of Evolved Skills.} 
The skill evolution pipeline may inherit biases or  unsafe heuristics from the trajectories and feedback used during optimization. 
Evolved skill libraries should therefore be inspected and validated before deployment.


\bibliography{custom}

\appendix

\section{Related Work}
\label{sec:related}

\paragraph{LLM agents and skill libraries.}
Equipping LLM agents with reusable procedural knowledge is a scalable approach to improve agent performance without modifying model weights.
Early efforts such as ReAct~\cite{yao2023react} and Reflexion~\cite{shinn2023reflexion} leverage textual feedback as in-context skill;
Voyager~\cite{wang2023voyager} maintains a growing skill library for Minecraft;
JARVIS-1~\cite{wang2023jarvis} and Ghost-in-the-Minecraft~\cite{zhu2023ghost} cache successful behaviors for replay at inference;
AgentTrek~\cite{xu2025agenttrek} bootstraps web agents with synthesized trajectories;
AutoManual~\cite{wang2024automanual} induces an \emph{operating manual} from interaction traces;
and ExpeL~\cite{zhao2024expel} distills cross-trial experiences into a reusable insights library.
More recent systems further elevate skills into first-class agent components:
SkillRL~\cite{xia2026skillrl} distills trajectories into a hierarchical SkillBank and recursively evolves skills with the agent policy;
and SkillOS~\cite{ouyang2026skillos} learns a long-horizon curator that inserts, updates, and deletes skills in an external SkillRepo.
Notably, SkVM~\cite{chen2026skvm} also identifies the model-skill mismatch problem---reporting that 87\% of tasks have at least one LLM that gains no benefit from the same skill---and addresses it by compiling skills into optimized runtime formats (e.g., code solidification, parallelization) to reduce latency and token cost.
\modelname{} shares the same motivation but pursues a complementary direction: rather than compiling skills for execution efficiency, we \emph{rewrite the natural-language expression} of skills to match each backbone's comprehension and reasoning style, directly improving task success rate.

\paragraph{Model-aware adaptation and prompt optimization.}
LLM behavior is highly sensitive to instruction phrasing even under semantically equivalent prompts~\cite{sclar2024quantifying}, motivating methods that tailor prompts to specific backbones.
Teacher-driven search methods such as OPRO~\cite{yang2024large} and EvoPrompt~\cite{guo2024connecting} iteratively refine a single instruction for a given task, yet treat the target model as fixed context---the same output applies regardless of backbone.
MAPO~\cite{chen2023mapo} and PromptBridge~\cite{wang2025promptbridge} further account for model identity by optimizing or transferring individual task instructions across backbones, yet they operate on single monolithic prompts in non-agent settings rather than on retrievable multi-entry skill libraries used at agent decision time.
\modelname{} differs in two key respects: 
(i) the optimization target is a dynamically retrieved \emph{skill library} rather than a monolithic prompt, 
and (ii) the evolutionary search is jointly steered by a structured model capability profile and environment reward signals, explicitly conditioning skill expression on target-model characteristics.
Furthermore, we train a lightweight skill rewriter that amortizes the expensive search process into a single forward pass---conceptually related to distilling costly inference-time computation into efficient learned models~\cite{singh2023beyond}---enabling skill adaptation to new backbones without repeated search.

\section{Ablations}
\label{sec:exp:ablation}

\paragraph{Two-stage evolution pipeline (Table~\ref{tab:ablation:search}).}
We ablate the two-stage search structure by replacing each stage's evolved skills with one-shot teacher (DeepSeek-V4) rewrites, isolating the contribution of each search stage.
\emph{w/o Task-specific} retains MASA-evolved general skills but substitutes teacher-written task-specific skills (i.e., Stage~1 only), while \emph{w/o General} retains MASA-evolved task-specific skills but uses teacher-written general skills (i.e., Stage~2 only).
Both stages contribute to the full pipeline, but their relative importance is environment- and scale-dependent.
On ALFWorld, removing task-specific search causes the largest drops for Qwen3-8B ($-25.0$) and Qwen3-32B ($-15.7$), indicating that per-task-type procedural guidance is critical for these backbones.
Conversely, removing general search most severely affects Qwen3-14B ($-16.4$), suggesting that high-level behavioral priors are essential when the model has sufficient capacity to follow them but still benefits from strategic framing.
On WebShop, removing general skills is catastrophic for 8B/14B/32B (SR drops to single digits), while removing task-specific skills has a comparatively modest effect.
This asymmetry reflects the nature of each environment---WebShop demands consistent high-level decision strategies that general skills encode, whereas ALFWorld requires fine-grained procedural sequences that task-specific skills address.

\paragraph{Rewriter model card (Table~\ref{tab:ablation:rewriter}).}
We ablate the model card input to the \modelname{}-Rewriter by comparing performance with and without the target model's capability card, using both training data variants: \emph{Cross-env} (trained on Search + WebShop) and \emph{Cross-task} (trained on Search + WebShop + ALFWorld Pick/Look/Pick2).
All results are average SR on three held-out ALFWorld tasks (Clean/Heat/Cool).
Removing the model card consistently degrades performance.
For Cross-task, the gap is especially pronounced on Qwen3-14B, confirming that the card provides critical conditioning signal for smaller backbones.
For Cross-env, model card removal also causes substantial drops.
Overall, these results suggest that the model card provides useful backbone-specific conditioning signals that help the rewriter generate more appropriate skill adaptations.

\begin{table}[H]
\centering
\small
\setlength{\tabcolsep}{4pt}
\begin{subtable}[t]{\columnwidth}
\centering
\begin{tabular}{lcccc}
\toprule
\textbf{Variant} & \textbf{4B} & \textbf{8B} & \textbf{14B} & \textbf{32B} \\
\midrule
\multicolumn{5}{l}{\textit{ALFWorld}} \\
Full pipeline              & \textbf{31.4} & \textbf{57.9} & \textbf{64.3} & \textbf{65.7} \\
\quad w/o Task-specific    & 25.0 & 32.9 & 63.6 & 50.0 \\
\quad w/o General          & 25.0 & 50.0 & 47.9 & 64.3 \\
\midrule
\multicolumn{5}{l}{\textit{WebShop}} \\
Full pipeline              & \textbf{26.4} & \textbf{28.6} & \textbf{29.2} & \textbf{34.6} \\
\quad w/o Task-specific    & 22.4 & 25.6 & 24.2 & 31.8 \\
\quad w/o General          & 23.4 & 7.2  & 10.2 & 9.6  \\
\bottomrule
\end{tabular}
\caption{The search-based evolution pipeline. }
\label{tab:ablation:search}
\end{subtable}

\vspace{0.5em}

\begin{subtable}[t]{\columnwidth}
\centering
\begin{tabular}{lcccc}
\toprule
\textbf{Variant} & \textbf{4B} & \textbf{8B} & \textbf{14B} & \textbf{32B} \\
\midrule
\multicolumn{5}{l}{\textit{Cross-env}} \\
\quad w/ Model Card        & \textbf{32.4} & \textbf{32.4} & \textbf{38.2} & \textbf{51.5} \\
\quad w/o Model Card       & 14.7 & 23.5 & 33.8 & 39.7 \\
\midrule
\multicolumn{5}{l}{\textit{Cross-task}} \\
\quad w/ Model Card        & \textbf{32.5} & \textbf{38.2} & \textbf{48.5} & \textbf{55.9} \\
\quad w/o Model Card       & 8.8  & 30.9 & 20.6 & 42.5 \\
\bottomrule
\end{tabular}
\caption{Model card conditioning in the \modelname{}-Rewriter.}
\label{tab:ablation:rewriter}
\end{subtable}

\caption{Ablation studies of \modelname{}.}
\label{tab:ablation}
\end{table}

\section{Preliminary Study: Supplementary Details}
\label{app:prelim_supp}

\subsection{Skill Variant Comparison}
\label{app:skill_variant_comparison}

Table~\ref{tab:app:preliminary_skills} shows concrete examples of the three non-empty ALFWorld skill variants used in the preliminary study. All variants keep the same skill IDs and task coverage; what changes is how much procedural text is exposed to the agent. We use bold text to highlight trigger conditions, executable steps, and failure-prevention cues added by the more detailed variants.

\begin{table*}[t]
\centering
\small
\setlength{\tabcolsep}{6pt}
\renewcommand{\arraystretch}{1.25}

\begin{subtable}{\textwidth}
\centering
\caption{General Skill: \emph{Systematic Exploration}}
\vspace{2pt}
\begin{tabular}{p{0.14\textwidth} p{0.78\textwidth}}
\toprule
\textbf{Granularity} & \textbf{Skill Text} \\
\midrule

\textsc{Concise} &
\textbf{Principle:}
Search all surfaces and containers once before revisiting.
\\

\cmidrule(lr){1-2}

\textsc{Moderate} &
\textbf{Principle:}
Search every plausible surface or container exactly once before revisiting; prioritize unopened or unseen locations to cover the whole room methodically.

\vspace{2pt}

\textbf{When to apply:}
Anytime the goal object count is not yet met and unexplored locations remain.
\\

\cmidrule(lr){1-2}

\textsc{Detailed} &
\textbf{Principle:}
When searching for an object, follow these steps exactly: \newline
    Step 1: Make a mental list of ALL possible locations in the room
    (countertop 1, countertop 2, shelf 1, drawer 1, cabinet 1, fridge 1, etc.). \newline
    Step 2: Visit each location one by one using 'go to [location] [number]'
    (e.g., 'go to countertop 1'). \newline
    Step 3: For closed containers (drawer, cabinet, fridge, safe, microwave),
    always use 'open [container] [number]' to check inside. \newline
    Step 4: Read the observation carefully — look for the exact name of the
    target object. \newline
    Step 5: Mark each location as 'checked' mentally and do NOT go back to it. \newline
    Step 6: Only after checking ALL locations in the room, consider that the
    object may not be present. \newline
    EXAMPLE: Looking for a mug → 'go to countertop 1' → check → 'go to
    countertop 2' → check → 'go to shelf 1' → check → 'open cabinet 1' →
    check inside → continue until found.

\textbf{When to apply:}
At the VERY START of every task that involves finding or
    locating any object. This is always your first action — never skip the
    systematic search.
\\

\bottomrule
\end{tabular}
\end{subtable}

\vspace{8pt}

\begin{subtable}{\textwidth}
\centering
\caption{Task-Specific Skill: \emph{Open Then Heat}}
\vspace{2pt}
\begin{tabular}{p{0.14\textwidth} p{0.78\textwidth}}
\toprule
\textbf{Granularity} & \textbf{Skill Text} \\
\midrule

\textsc{Concise} &
\textbf{Principle:}
Open microwave, put object in, heat it.
\\

\cmidrule(lr){1-2}

\textsc{Moderate} &
\textbf{Principle:}
Upon reaching the microwave with the target in hand, always open the door, place the object inside, and execute the heat action before leaving.

\vspace{2pt}

\textbf{When to apply:}
Immediately after navigating to the microwave with the target
    object held.
\\

\cmidrule(lr){1-2}

\textsc{Detailed} &
\textbf{Principle:}
The microwave heating sequence must be executed in this EXACT order:  \newline
    (1) 'go to microwave 1' — navigate to the microwave.  \newline
    (2) 'open microwave 1' — the door must be open to put things in. \newline
    (3) 'put [object] in/on microwave 1' — place the object inside. \newline
    (4) 'heat [object] with microwave 1' — execute the heating action. \newline
    (5) 'open microwave 1' — open the door again to retrieve (if needed). \newline
    (6) 'take [object] from microwave 1' — take the now-heated object. \newline
    COMMON MISTAKE: Trying to 'heat' without first putting the object in the
    microwave → fails. \newline
    ANOTHER MISTAKE: Forgetting to open the microwave before putting the object
    in → fails.

\vspace{2pt}

\textbf{When to apply:}
When you are holding the target object and ready to heat it.  Execute this exact 6-step sequence.
\\

\bottomrule
\end{tabular}
\end{subtable}

\caption{
Examples from the ALFWorld skill-library variants used in the preliminary study.
We show the same cross-task skill and task-specific skill under three granularity levels:
\textsc{Concise}, \textsc{Moderate}, and \textsc{Detailed}.
The empty-bank control (\textsc{No Skill}) is omitted for brevity.
}
\label{tab:app:preliminary_skills}

\end{table*}

\subsection{Per-Task Breakdown: Qwen3}
\label{app:prelim_per_task}

Table~\ref{tab:app:prelim_per_task} expands the overall numbers visualized in Figure~\ref{fig:pre_qwen} into per-task success rates for each (model, skill) cell. The breakdown is computed on the same ALFWorld validation set. Note the large within-condition swings across task types (e.g., Qwen3-14B Concise: $74.2$ on \textsc{Pick} vs.\ $13.7$ on \textsc{Cool}; Qwen3-4B Detailed: $1.6$ on \textsc{Pick} vs.\ $46.7$ on \textsc{Look}), which are substantially larger than cross-condition differences and motivate the task-specific tree-search stage of the evolution pipeline.

\begin{table*}[t]
\centering
\small
\setlength{\tabcolsep}{4pt}
\begin{tabular}{llccccccc}
\toprule
\textbf{Model} & \textbf{Skill} & \textbf{Pick} & \textbf{Clean} & \textbf{Heat} & \textbf{Cool} & \textbf{Pick2} & \textbf{Look} & \textbf{Overall} \\
\midrule
\multirow{4}{*}{Qwen3-4B}
 & No Skill   & 20.0 & 18.5 & 18.8 & 16.0 & 12.5 & 15.4 & 17.1 \\
 & Concise    & 3.2  & 17.9 & 37.5 & 9.1  & 10.7 & 40.0 & 16.4 \\
 & Moderate   & 20.0 & 29.6 & 12.5 & 20.0 & 8.3  & 30.8 & \textbf{20.0} \\
 & Detailed   & 1.6  & 16.1 & 9.3  & 6.8  & 10.7 & 46.7 & 12.8 \\
\midrule
\multirow{4}{*}{Qwen3-8B}
 & No Skill   & 54.3 & 29.6 & 6.2  & 24.0 & 20.8 & 46.2 & \textbf{32.1} \\
 & Concise    & 32.2 & 33.9 & 25.0 & 11.3 & 25.0 & 40.0 & 27.9 \\
 & Moderate   & 17.1 & 40.7 & 31.2 & 20.0 & 12.5 & 38.5 & 25.0 \\
 & Detailed   & 6.5  & 17.9 & 21.9 & 13.6 & 10.7 & 50.0 & 17.1 \\
\midrule
\multirow{4}{*}{Qwen3-14B}
 & No Skill   & 65.7 & 25.9 & 43.8 & 16.0 & 29.2 & 38.5 & 37.9 \\
 & Concise    & 74.2 & 26.8 & 18.8 & 13.7 & 30.3 & 43.4 & 36.8 \\
 & Moderate   & 68.6 & 44.4 & 25.0 & 20.0 & 33.3 & 46.2 & 42.1 \\
 & Detailed   & 64.5 & 46.4 & 18.8 & 34.1 & 46.4 & 66.7 & \textbf{47.5} \\
\midrule
\multirow{4}{*}{Qwen3-32B}
 & No Skill   & 48.6 & 44.4 & 25.0 & 32.0 & 16.7 & 46.2 & 36.4 \\
 & Concise    & 56.5 & 41.0 & 37.5 & 18.2 & 33.9 & 56.6 & 40.7 \\
 & Moderate   & 48.6 & 40.7 & 50.0 & 44.0 & 20.8 & 46.2 & 41.4 \\
 & Detailed   & 54.8 & 46.4 & 31.2 & 36.4 & 28.6 & 60.0 & \textbf{42.9} \\
\bottomrule
\end{tabular}
\caption{Per-task ALFWorld success rate (\%) for the four Qwen3 backbones under each skill granularity condition.}
\label{tab:app:prelim_per_task}
\end{table*}

\subsection{Supplementary Validation: Gemma3}
\label{app:cross_family}

\begin{figure}[t]
    \centering
    \includegraphics[width=0.95\linewidth]{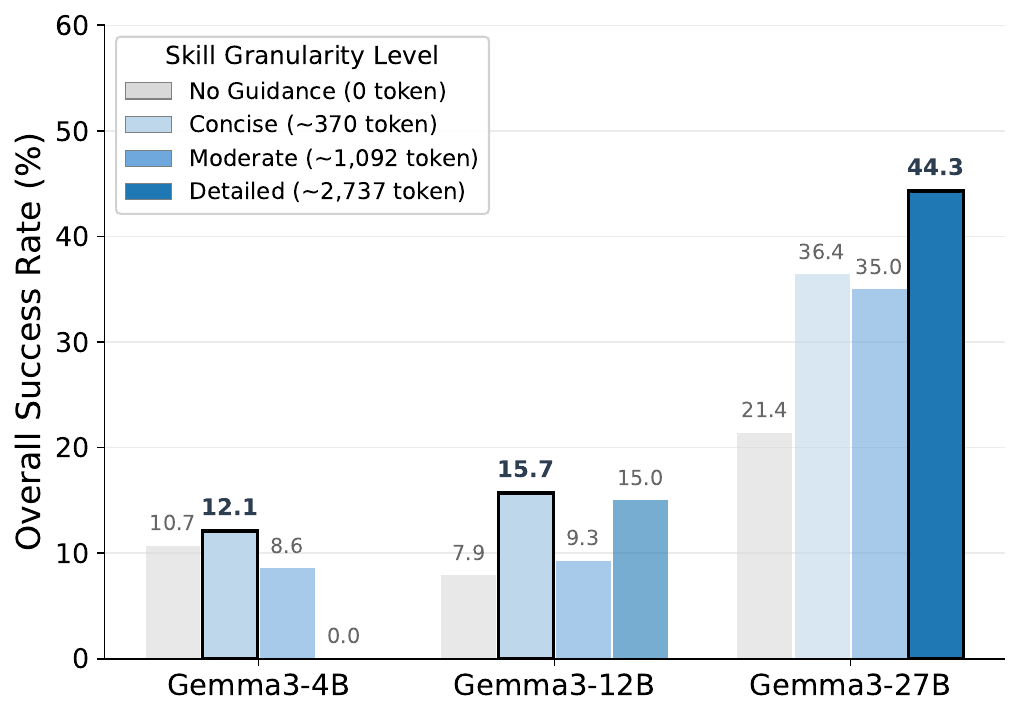}
    \caption{The supplementary validation of Gemma family.
    }
    \label{fig:app:gemma_verbosity}
\end{figure}

To verify that the scale-dependent granularity pattern is not unique to Qwen3, we repeat the fixed-granularity sweep on Gemma3 backbones (4B/12B/27B)~\cite{kamath2025gemma3}. 
It supports the motivating conclusion: the best skill form is model-dependent rather than universally transferable. 
Gemma3-4B and Gemma3-12B are strongest with Concise skills, while Gemma3-27B reaches its best success rate with Detailed skills. Figure~\ref{fig:app:gemma_verbosity} shows the overall results and Table~\ref{tab:app:gemma_per_task} gives the per-task breakdown.

Comparing models of the same parameter count across families
further isolates the effect of architecture and training from that of scale alone.
Gemma3-4B achieves its best performance with \texttt{Concise}
  skills, whereas Qwen3-4B peaks under \texttt{Moderate} skills
  (Figure~\ref{fig:pre_qwen} vs.\ Figure~\ref{fig:app:gemma_verbosity}).
  Despite identical parameter budgets, the two models respond to
  skill granularity in qualitatively different ways—confirming that
  the optimal skill form is determined by a model's overall
  characteristics (architecture, training data, alignment procedure)
  rather than parameter count alone.
This observation reinforces the necessity of conditioning skill
adaptation on a rich model profile rather than relying on scale as a proxy.

\begin{table*}[t]
\centering
\small
\setlength{\tabcolsep}{4pt}
\begin{tabular}{llccccccc}
\toprule
\textbf{Model} & \textbf{Skill} & \textbf{Pick} & \textbf{Clean} & \textbf{Heat} & \textbf{Cool} & \textbf{Pick2} & \textbf{Look} & \textbf{Overall} \\
\midrule
\multirow{4}{*}{Gemma3-4B}
 & No Skill & 22.6 & 0.0 & 0.0 & 0.0 & 7.1 & 40.0 & 10.7 \\
 & Concise  & 22.6 & 7.1 & 6.2 & 0.0 & 7.1 & 33.3 & \textbf{12.1} \\
 & Moderate & 3.2 & 7.1 & 0.0 & 4.5 & 7.1 & 40.0 & 8.6 \\
 & Detailed & 0.0 & 0.0 & 0.0 & 0.0 & 0.0 & 0.0 & 0.0 \\
\midrule
\multirow{4}{*}{Gemma3-12B}
 & No Skill & 16.1 & 3.6 & 0.0 & 0.0 & 3.6 & 26.7 & 7.9 \\
 & Concise  & 32.3 & 10.7 & 12.5 & 0.0 & 7.1 & 33.3 & \textbf{15.7} \\
 & Moderate & 9.7 & 10.7 & 0.0 & 0.0 & 7.1 & 33.3 & 9.3 \\
 & Detailed & 38.7 & 21.4 & 6.2 & 0.0 & 3.6 & 6.7 & 15.0 \\
\midrule
\multirow{4}{*}{Gemma3-27B}
 & No Skill & 22.6 & 14.3 & 18.8 & 13.6 & 25.0 & 40.0 & 21.4 \\
 & Concise  & 61.3 & 21.4 & 31.2 & 9.1 & 42.9 & 46.7 & 36.4 \\
 & Moderate & 51.6 & 32.1 & 18.8 & 4.5 & 42.9 & 53.3 & 35.0 \\
 & Detailed & 41.9 & 60.7 & 31.2 & 31.8 & 35.7 & 66.7 & \textbf{44.3} \\
\bottomrule
\end{tabular}
\caption{Per-task ALFWorld success rate (\%) for the three Gemma3 backbones under each skill granularity condition.}
\label{tab:app:gemma_per_task}
\end{table*}

\section{Model Card Construction}
\label{app:model_cards}

Each model card is constructed from a fixed rubric combining public documentation and automated analysis:
\begin{enumerate}[leftmargin=*]
\item \emph{Architecture metadata.} Model family, variant name, parameter count, architecture type, layer/attention configuration, context window, and vocabulary size---sourced directly from the published model card or config files.
\item \emph{Training provenance.} Whether the checkpoint is base or instruction-tuned, the alignment pipeline (e.g., SFT + DPO + GRPO), training data scale, and multilingual support---sourced from official documentation.
\item \emph{Capability profile.} Strengths are extracted from the model's official release notes (e.g., ``strong at math and code generation''). Weaknesses are generated by the teacher LLM summarizing behavioral patterns observed during a small set of preliminary rollouts (Section~\ref{sec:prelim}), produced automatically without human annotation.
\end{enumerate}

\noindent Note that the card does not include any downstream evaluation results (e.g., ALFWorld success rates) or oracle style labels (e.g., \emph{prefers\_concise}). The preliminary rollouts used for weakness summarization are disjoint from the evaluation set. 

Below is the card for Qwen3-4B; cards for the remaining backbones follow the same template.

\begin{lstlisting}[basicstyle=\ttfamily\small,frame=single,breaklines=true,label=lst:card_4b]
# Model Card: Qwen3-4B
# Source: https://huggingface.co/Qwen/Qwen3-4B

# === Architecture Metadata ===
family: "Qwen3"
variant: "4B"
parameter_count: "4B"
architecture: "dense-transformer"
num_layers: 36
hidden_size: 2560
num_attention_heads: 32
num_kv_heads: 8
context_window: 32768
vocab_size: 151936

# === Training Provenance ===
base_or_instruct: "instruct"
alignment_method: "SFT + DPO + GRPO"
training_data_size: "36T tokens"
multilingual: true

# === Official Capabilities (from release notes) ===
strengths: "math, code generation, instruction
  following, multilingual, tool use, thinking
  mode support"

# === Observed Weaknesses (teacher-summarized) ===
weaknesses: "limited reasoning depth due to small
  parameter count, may struggle with complex
  multi-step planning"

\end{lstlisting}

\section{Skill Evolution Pipeline Details}
\label{app:hyperparams}

This appendix provides the full algorithmic procedures (Algorithms~\ref{alg:stage1} and~\ref{alg:stage2}) and hyperparameters for the two-stage skill evolution pipeline described in Section~\ref{sec:method:rsep}.
All evolution experiments use the original training split of each environment for exploration; the evaluation results reported in the main paper are obtained on the held-out test split.

\begin{algorithm}[t]
\small
\caption{Stage 1: General Skill Search (Hill Climbing)}
\label{alg:stage1}
\begin{algorithmic}[1]
\REQUIRE Target model $F$, model card $\mathcal{M}_F$, teacher $T$, eval set $\mathcal{D}$ (sampled from training episodes), initial general skills $\mathcal{S}^{G_0}_F$, max iterations $I$, patience $p$, history size $K$
\ENSURE Optimized general skills $\mathcal{S}^{G\star}_F$
\STATE $\mathcal{S}^{G\star}_F \gets \mathcal{S}^{G_0}_F$ \COMMENT{current best general skill set}
\STATE $R^\star \gets \mathrm{Eval}(F, \mathcal{S}^{G\star}_F, \mathcal{D})$ \COMMENT{its average reward across all task types}
\STATE $\mathcal{H} \gets \{(\mathcal{S}^{G_0}_F, R^\star)\}$ \COMMENT{search history: (skill set, reward) pairs}
\FOR{$i = 1$ \TO $I$}
  \STATE \textcolor{gray}{\textit{// Rollout \& Analysis}}
  \STATE $\mathcal{F}_i \gets \mathrm{CollectFailures}(F, \mathcal{S}^{G\star}_F, \mathcal{D})$
  \STATE $\mathrm{attr}_i \gets T.\mathrm{Analyze}(\mathcal{F}_i)$ \COMMENT{structured failure attribution}
  \STATE \textcolor{gray}{\textit{// Rewrite}}
  \STATE $\mathcal{S}^{G_i}_F \gets T.\mathrm{Rewrite}(\mathcal{S}^{G\star}_F, \mathrm{attr}_i, \mathrm{TopK}(\mathcal{H}, K), \mathcal{M}_F)$
  \STATE \textcolor{gray}{\textit{// Accept / Reject}}
  \STATE $R_i \gets \mathrm{Eval}(F, \mathcal{S}^{G_i}_F, \mathcal{D})$
  \STATE $\mathcal{H} \gets \mathcal{H} \cup \{(\mathcal{S}^{G_i}_F, R_i)\}$
  \IF{$R_i > R^\star$}
    \STATE $\mathcal{S}^{G\star}_F \gets \mathcal{S}^{G_i}_F$;\; $R^\star \gets R_i$ \COMMENT{accept}
  \ENDIF
  \IF{no improvement for $p$ consecutive iterations}
    \STATE \textbf{break}
  \ENDIF
\ENDFOR
\RETURN $\mathcal{S}^{G\star}_F$
\end{algorithmic}
\end{algorithm}

\begin{algorithm}[t]
\small
\caption{Stage 2: Task-Specific Skill Search (Per-Type Tree Search)}
\label{alg:stage2}
\begin{algorithmic}[1]
\REQUIRE Target model $F$, model card $\mathcal{M}_F$, teacher $T$, fixed general skills $\mathcal{S}^{G\star}_F$, initial task-specific skills $\{\mathcal{S}^{T_{c_0}}_F\}_{c \in \mathcal{C}}$, iterations $J$
\ENSURE Optimized task-specific skills $\{\mathcal{S}^{T_c\star}_F\}_{c \in \mathcal{C}}$
\FOR{each task type $c \in \mathcal{C}$ \textbf{in parallel}}
  \STATE Initialize tree root with $\mathcal{S}^{T_{c_0}}_F$
  \FOR{$j = 1$ \TO $J$}
    \STATE \textcolor{gray}{\textit{// Selection}}
    \STATE $n \gets \mathrm{UCB1Select}(\text{root})$ \COMMENT{select leaf via Eq.~\ref{eq:ucb1}}
    \STATE \textcolor{gray}{\textit{// Expansion}}
    \STATE $\mathcal{F} \gets \mathrm{CollectFailures}(F, \mathcal{S}^{G\star}_F, \mathcal{S}^{T_c}_{F,n}, c)$
    \STATE $\mathrm{attr} \gets T.\mathrm{Analyze}(\mathcal{F})$ \COMMENT{failure attribution}
    \STATE $\mathcal{S}'^{T_c}_F \gets T.\mathrm{Rewrite}(\mathcal{S}^{T_c}_{F,n},\, \mathrm{attr},\, \mathcal{M}_F)$
    \STATE Add $\mathcal{S}'^{T_c}_F$ as child of node $n$
    \STATE \textcolor{gray}{\textit{// Evaluation}}
    \STATE $R' \gets \mathrm{Eval}(F, \mathcal{S}^{G\star}_F, \mathcal{S}'^{T_c}_F, c)$
    \STATE \textcolor{gray}{\textit{// Backpropagation}}
    \STATE Update visit counts and value estimates from new node to root
  \ENDFOR
  \STATE $\mathcal{S}^{T_c\star}_F \gets$ skill set of the highest-value node
\ENDFOR
\RETURN $\{\mathcal{S}^{T_c\star}_F\}_{c \in \mathcal{C}}$
\end{algorithmic}
\end{algorithm}

\subsection{Stage 1: Hill Climbing}

Maximum iterations $I{=}10$; patience $p{=}3$ (early stopping after 3 consecutive iterations without improvement); top-$K{=}5$ highest-reward historical skill sets provided to the teacher at each iteration.
A candidate general skill set is accepted if and only if its average adjusted reward strictly exceeds the current best.

\subsection{Stage 2: UCB-Driven Tree Search}

At each iteration, the node $n$ maximizing the following UCB1 score is selected:
\begin{equation}
\label{eq:ucb1}
\mathrm{UCB1}(n) = \bar{R}(n) + C \sqrt{\frac{\ln N_{\mathrm{parent}}}{N_n}},
\end{equation}
where $\bar{R}(n)$ is the mean adjusted reward of node $n$ and all its descendants, $N_n$ is the visit count of node $n$, $N_{\mathrm{parent}}$ is the visit count of its parent, and $C{=}1.4$ is the exploration constant.
We run $J{=}10$ iterations per task type with $N{=}100$ episodes per node evaluation.

\section{Skill Rewriter Training Details}
\label{app:mcsr_train}

We perform full-parameter SFT on Qwen3-4B in BF16 precision.
Training uses AdamW (lr $1\mathrm{e}{-5}$, cosine schedule, warmup ratio $0.1$, gradient checkpointing), effective batch size $4$ (per-device $1 \times$ gradient accumulation $4$), $5$ epochs, and max sequence length $4096$.
We select the best checkpoint based on training loss convergence.

The training data consists of pairs for in-domain tasks, with data augmentation including noisy inputs (noise ratio $0.3$), partial inputs (keep ratio $0.6$), and cross-model transfer pairs.
We train two rewriter variants: a combined rewriter on 769 samples from all three environments (ALFWorld Pick/Look/Pick2 only---excluding the held-out types, WebShop, and Search), and an environment-specific rewriter on 499 samples (WebShop + Search only).

\section{WebShop Supplementary Results}
\label{app:webshop_trajectory}

\subsection{Trajectory Analysis: Why Larger Models Fail}

We analyze failed WebShop trajectories to understand why larger Qwen3 models (8B/14B/32B) perform worse than 4B under baseline conditions (Section~\ref{sec:exp:main}).

Table~\ref{tab:webshop_trajectory} reveals a striking pattern: Qwen3-4B produces concise, action-only outputs (0\% steps with chain-of-thought, ${\sim}73$ chars per action), while 8B/14B/32B prepend extensive reasoning preambles before each action command.
Qwen3-14B is the most severe case, with 97\% of steps containing verbose reasoning.
This behavior exhausts the fixed step budget on deliberation rather than environment interaction---the agent ``thinks'' through multiple options but never completes enough purchase actions to succeed.

\begin{table}[H]
\centering
\small
\setlength{\tabcolsep}{8pt}
\renewcommand{\arraystretch}{1.05}
\begin{tabular}{l r r}
\toprule
\textbf{Model} & \textbf{CoT (\%)} & \textbf{Action Len.} \\
\midrule
Qwen3-4B  & 0  & 73 chars  \\
Qwen3-8B  & 57 & 1,021 chars \\
Qwen3-14B & 97 & 574 chars  \\
Qwen3-32B & 66 & 491 chars  \\
\bottomrule
\end{tabular}
\caption{WebShop trajectory statistics. CoT: fraction of steps containing reasoning preambles.}
\label{tab:webshop_trajectory}
\end{table}

\subsection{Per-Category Breakdown}

Table~\ref{tab:webshop_per_category} provides the full per-category success rate breakdown.
Several observations stand out:
\begin{itemize}[leftmargin=*]
\item For 8B/14B/32B baselines, most categories have near-zero SR, consistent with the verbose-reasoning bottleneck identified above.
\item \modelname{} achieves the best SR in the vast majority of categories across all backbones, with particularly large gains on \textsc{Other} and \textsc{Electronics}.
\item The improvement is broad rather than category-specific: \modelname{} does not exploit a single easy category to inflate the average but improves performance across the board.
\end{itemize}

\begin{table*}[t]
\centering
\small
\setlength{\tabcolsep}{4pt}
\renewcommand{\arraystretch}{1.08}
\begin{tabular}{ll rrrrrrr | r}
\toprule
& & \multicolumn{7}{c|}{\textbf{Per-Category SR (\%)}} & \\
\cmidrule(lr){3-9}
\textbf{Model} & \textbf{Method}
& \textbf{Apparel}
& \textbf{Other}
& \textbf{Footwear}
& \textbf{Home}
& \textbf{Elec.}
& \textbf{Access.}
& \textbf{Beauty}
& \textbf{Avg.} \\
\midrule

\multirow{4}{*}{Qwen3-4B}
& No Skill       & 18.6 & 40.0 & 11.1 & 19.0 & 71.4 & 20.0 & 16.7 & 23.0 \\
& + Base Skill   & 16.1 & 27.0 & 26.7 & 19.0 & 28.6 & 10.0 & 16.7 & 19.4 \\
& + DS-Adapter   & 20.3 & 20.0 & 13.3 & 14.3 & 28.6 & 10.0 & 16.7 & 19.2 \\
\rowcolor{masahighlight}
& + \modelname{} & \textbf{23.2} & \textbf{38.0} & \textbf{28.9} & \textbf{19.0} & 28.6 & \textbf{20.0} & \textbf{16.7} & \textbf{26.4} \\
\midrule

\multirow{4}{*}{Qwen3-8B}
& No Skill       &  2.6 & 11.0 &  0.0 &  4.8 & 14.3 & \textbf{20.0} &  0.0 &  4.6 \\
& + Base Skill   &  5.5 &  9.0 &  0.0 &  0.0 & 14.3 & \textbf{20.0} & 16.7 &  6.0 \\
& + DS-Adapter   &  4.8 &  5.0 &  0.0 &  0.0 &  0.0 & 10.0 & 16.7 &  4.4 \\
\rowcolor{masahighlight}
& + \modelname{} & \textbf{27.7} & \textbf{41.0} & \textbf{13.3} & \textbf{14.3} & \textbf{57.1} & 10.0 & \textbf{33.3} & \textbf{28.6} \\
\midrule

\multirow{4}{*}{Qwen3-14B}
& No Skill       &  1.0 &  6.0 &  6.7 &  4.8 &  0.0 & 10.0 &  0.0 &  2.8 \\
& + Base Skill   &  0.3 &  2.0 &  4.4 &  4.8 &  0.0 & 20.0 &  0.0 &  1.6 \\
& + DS-Adapter   &  0.3 &  6.0 &  2.2 &  4.8 &  0.0 & 10.0 &  0.0 &  2.0 \\
\rowcolor{masahighlight}
& + \modelname{} & \textbf{32.8} & \textbf{33.0} & \textbf{6.7} & \textbf{9.5} & \textbf{28.6} & \textbf{30.0} & \textbf{16.7} & \textbf{29.2} \\
\midrule

\multirow{4}{*}{Qwen3-32B}
& No Skill       &  4.2 & 16.0 &  2.2 &  9.5 &  0.0 & 10.0 &  0.0 &  6.6 \\
& + Base Skill   &  4.8 & 14.0 &  4.4 &  9.5 & 14.3 & 20.0 &  0.0 &  7.2 \\
& + DS-Adapter   &  2.3 &  8.0 &  2.2 &  4.8 &  0.0 & 10.0 &  0.0 &  3.6 \\
\rowcolor{masahighlight}
& + \modelname{} & \textbf{32.2} & \textbf{48.0} & \textbf{26.7} & \textbf{19.0} & \textbf{42.9} & \textbf{40.0} & \textbf{33.3} & \textbf{34.6} \\
\bottomrule
\end{tabular}
\caption{WebShop per-category success rate (\%). \textbf{Bold} marks the best within each backbone.}
\label{tab:webshop_per_category}
\end{table*}

\section{Skill Rewriter OOD: Per-Task Breakdown}
\label{app:mcsr_ood_pertask}

\begin{figure*}[t]
    \centering
    \includegraphics[width=\textwidth]{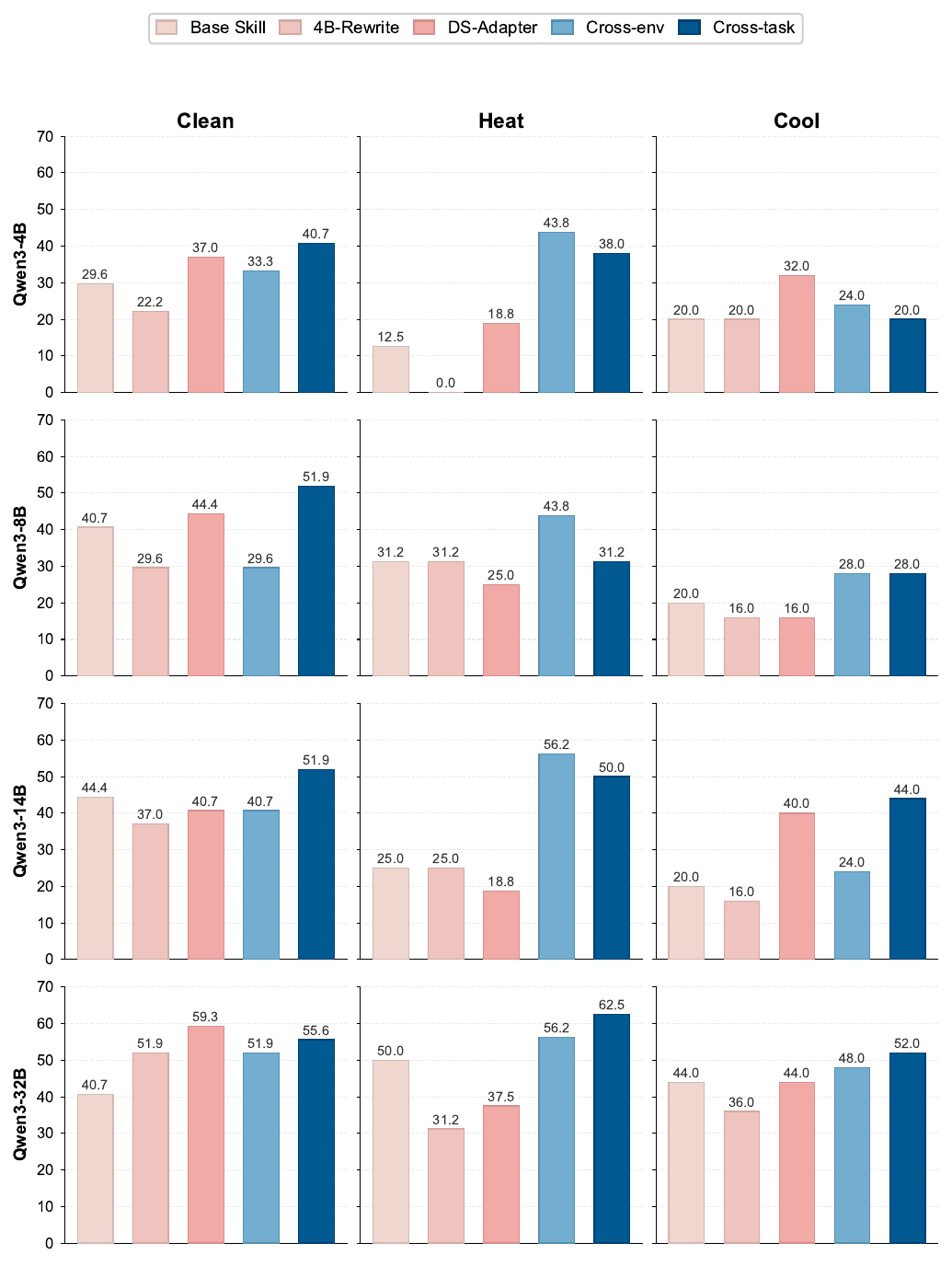}
    \caption{Per-task OOD generalization of \modelname{}-Rewriter. Rows: target backbones (4B--32B). Columns: held-out task types (Clean, Heat, Cool).}
    \label{fig:ood_4x3}
\end{figure*}

Figure~\ref{fig:ood_4x3} shows the per-task SR breakdown for the OOD generalization experiment (Section~\ref{sec:exp:mcsr_ood}).

\paragraph{Cross-task transfer.}
Adding ALFWorld \textsc{Pick}/\textsc{Look}/\textsc{Pick2} traces to the training set (dark blue) yields consistent improvements over Cross-env on all three tasks.
The gains are most pronounced on \textsc{Cool} (e.g., 14B: 24.0$\to$44.0; 32B: 48.0$\to$52.0) and \textsc{Clean} (e.g., 8B: 29.6$\to$51.9; 14B: 40.7$\to$51.9), indicating that in-environment traces help the rewriter learn ALFWorld-specific action patterns such as navigation sequences and object interaction protocols.
On \textsc{Heat}, Cross-task improves for 4B and 14B but is comparable to Cross-env for 8B and 32B, suggesting that Heat-specific patterns are already partially captured by the cross-environment signal.

\paragraph{Cross-environment transfer.}
Trained only on Search and WebShop traces, the Cross-env rewriter (light blue) shows notable strengths on \textsc{Heat} across all backbones---particularly 14B (56.2\%) and 32B (56.2\%)---substantially exceeding DS-Adapter.
On \textsc{Clean} and \textsc{Cool}, Cross-env performance is more mixed: it matches or slightly exceeds DS-Adapter for 4B and 8B, but falls short on some larger-backbone cells (e.g., 14B \textsc{Clean}: 40.7 vs.\ DS-Adapter 40.7, tied).
This suggests that cross-environment transfer is most effective when the target task involves decision patterns (e.g., sequential verification in \textsc{Heat}) that overlap with those in the training environments.

\section{Qualitative Analysis: Evolved Skill Examples}
\label{app:analysis}

\begin{table*}[t]
\centering
\small
\setlength{\tabcolsep}{6pt}
\renewcommand{\arraystretch}{1.2}
\begin{tabular}{p{0.04\textwidth} p{0.58\textwidth} p{0.14\textwidth} p{0.14\textwidth}}
\toprule
\textbf{Model} & \textbf{Evolved Skill Text} & \textbf{Failure Mode} & \textbf{Strategy} \\
\midrule

4B &
\texttt{When selecting a color variant, match the EXACT string from the task requirement to the available options. `navy blue' $\neq$ `light blue' $\neq$ `navy'. `c3-black' $\neq$ `c-black'. If the exact color name is not available in the options list, this product CANNOT satisfy your requirement---\textbf{\textcolor{red}{leave immediately}}. \textbf{\textcolor{red}{Do NOT select an approximate or similar color.}}}
& Picks visually similar colors by guessing & Strict binary match: exact or leave \\

\cmidrule(lr){1-4}

8B &
\texttt{Scan ALL color options. If the EXACT color name appears, click it. `c3-black' is NOT the same as `a6-black'. If the exact color is NOT available but a SIMILAR one exists (e.g., goal says `green', options have `e-green'), \textbf{\textcolor{blue}{select the CLOSEST match and proceed to buy}}. A close color match gives partial credit which is better than 0. \textbf{\textcolor{blue}{Even if the product doesn't perfectly match---BUY IT.}}}
& Abandons products too easily (0 credit) & Flexible match: buy anyway for partial credit \\

\cmidrule(lr){1-4}

14B &
\texttt{If your required color is NOT in the admissible actions list, \textbf{\textcolor{teal}{click `back to search' immediately}}. Do not try similar colors. Do not try similar sizes. \textbf{\textcolor{teal}{One glance at options $\to$ if exact match missing $\to$ back to search. Takes 1 step, not 5.}}}
& Wastes steps deliberating on bad products & Fast-fail: 1-step exit if no exact match \\

\cmidrule(lr){1-4}

32B &
\texttt{\textbf{\textcolor{violet}{Match color by checking if the task's required color name appears as a SUBSTRING}} in any admissible action, or vice versa. `patina green' matches `patina green' (exact) but NOT `yellow'. `green' matches `a1-green' or `d01green' (contains). \textbf{\textcolor{violet}{When multiple options contain the color word, prefer the one that matches more of the full color name.}}}
& Mishandles coded names (e.g.\ \texttt{d01green}) & Algorithmic: substring matching with preference rule \\

\bottomrule
\end{tabular}
\caption{WebShop color-matching skill evolved by \modelname{} for four backbones. \textcolor{red}{Red}: rigid rejection rule (4B); \textcolor{blue}{Blue}: flexible buy-anyway heuristic (8B); \textcolor{teal}{Teal}: 1-step fast-fail exit (14B); \textcolor{violet}{Violet}: algorithmic substring matching (32B). Each strategy targets the dominant failure mode of its target model.}
\label{tab:color_matching}
\end{table*}

Table~\ref{tab:color_matching} presents a case study of how \modelname{} adapts skills differently for each backbone on the same subtask---WebShop's color-matching decision, identified as the highest-failure-rate subtask during skill evolution.

Rather than producing minor wording variations, the evolution pipeline discovers qualitatively distinct strategies tailored to each model's dominant failure mode:
\begin{itemize}[leftmargin=*]
\item Qwen3-4B tends to guess visually similar colors. The evolved skill imposes a strict binary rule: match exactly or leave immediately.
\item Qwen3-8B abandons products too easily, scoring zero. The evolved skill encourages buying approximate matches for partial credit.
\item Qwen3-14B wastes steps deliberating on bad products. The evolved skill enforces a one-step fast-fail exit when no exact match exists.
\item Qwen3-32B mishandles coded color names (e.g., \texttt{d01green}). The evolved skill provides an algorithmic substring-matching procedure with tie-breaking rules.
\end{itemize}

\noindent This demonstrates that model-conditioned adaptation operates at the level of \emph{decision strategy}---the same problem requires fundamentally different solutions depending on how each backbone fails.

\section{The Use of Large Language Models (LLMs)}

In this paper, large language models were utilized exclusively for grammatical polishing and stylistic refinement, aimed at enhancing the clarity and readability of our presentation of results.

\vspace{1em}

\textbf{\emph{
The following pages contain supplementary tables and  figures.}}

\end{document}